%% file: root.tex
\documentclass[letterpaper, 10 pt, conference]{ieeeconf}  %

\usepackage{times}
\usepackage{epsfig}
\usepackage{graphicx}
\usepackage{amsmath}
\usepackage{amssymb}
\usepackage{multirow}
\usepackage[linesnumbered,ruled]{algorithm2e}
\usepackage{bm}
\usepackage{dblfloatfix}

\IEEEoverridecommandlockouts                              %

\newcommand\blfootnote[1]{%
  \begingroup
  \renewcommand\thefootnote{}\footnote{#1}%
  \addtocounter{footnote}{-1}%
  \endgroup
}

\title{\LARGE \bf
Oriented Point Sampling for Plane Detection in Unorganized Point Clouds
}

\author{Bo Sun$^{1}$ and Philippos Mordohai$^{1}$%
\thanks{$^{1}$Bo Sun and Philippos Mordohai are with the Department of Computer Science, Stevens Institute of Technology, Hoboken, NJ 07030, USA
        ${\tt\small \left\{bsun7, Philippos.Mordohai\right\}@stevens.edu.}$}%
}

\begin{document}

\maketitle
\thispagestyle{empty}
\pagestyle{empty}

\begin{abstract}

\input{abstract}

\end{abstract}

\section{INTRODUCTION}
\input{introduction}

\section{RELATED WORK}

\input{relatedwork}

\section{APPROACH} \label{sec:approach}
\input{approach}

\section{EXPERIMENTS} \label{sec:experiments}
\input{experiment}

\section{CONCLUSIONS}
\input{conclusion}

\bibliographystyle{IEEEtran}
\bibliography{IEEEexample}

\end{document}

%% file: abstract.tex
Plane\blfootnote{This work was supported in part by the National Institute Of Nursing Research of the National Institutes of Health under Award R01NR015371 and the National Science Foundation under Award IIS-1637761.} detection in 3D point clouds is a crucial pre-processing step for applications such as point cloud segmentation, semantic mapping and SLAM. In contrast to many recent plane detection methods that are only applicable on organized point clouds, our work is targeted to unorganized point clouds that do not permit a 2D parametrization. We compare three methods for detecting planes in point clouds efficiently. One is a novel method proposed in this paper that generates plane hypotheses by sampling from a set of points with estimated normals. We named this method Oriented Point Sampling (OPS) to contrast with more conventional techniques that require the sampling of three unoriented points to generate plane hypotheses. We also implemented an efficient plane detection method based on local sampling of three unoriented points and compared it with OPS and the 3D-KHT algorithm, which is based on octrees, on the detection of planes on 10,000 point clouds from the SUN RGB-D dataset.

%% file: introduction.tex
As depth cameras and 3D sensors have become widely available recently, they are extensively used in various robotics and computer vision applications, such as Simultaneous Localization and Mapping (SLAM) \cite{kerl2013dense,whelan2013robust,ma2016cpa,yang2016pop}, dense 3D reconstruction \cite{nuchter2008towards,newcombe2011kinectfusion,henry2012rgb,niessner2013real,choi2015robust,klingensmith2015chisel,dzitsiuk2017noising} and 3D object recognition \cite{schnabel2007efficient,qian20143d}.
3D point clouds acquired by such depth cameras are generally noisy and redundant, and do not provide semantics of the scene.

An important first step to obtain more useful representations is to detect planar segments in the point clouds. In outdoor scenes, the ground is typically piecewise planar. In indoor scenes, most important surfaces, including ceilings, walls and floors,  are planar.
We can divide the planes into three groups: (1) horizontal planes, which are important due to their roles as support surfaces for other objects and as potentially traversable terrain for robots; (2) vertical planes, which are also important because many obstacles and the walls are vertical; (3) other planes, which constitute most of the other objects. It should be noted, however, that many of the recent plane detection approaches \cite{pop2008fast,biswas2011fast,biswas2012depth,feng2014fast,feng2014fast,marriott2017plane} accept organized point clouds, i.e. $2\frac{1}{2}$-D depth images, as input. Here, we are interested in developing methods applicable on unorganized point clouds. This requires the search for neighboring points in 3D \cite{muja2014scalable} since $2\frac{1}{2}$-D information cannot be exploited.

\begin{figure}
\begin{center}
\begin{tabular}{cc}
\includegraphics[width=.22 \textwidth]{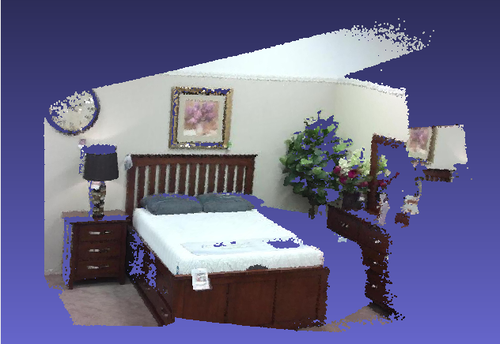} & \includegraphics[width=.22 \textwidth]{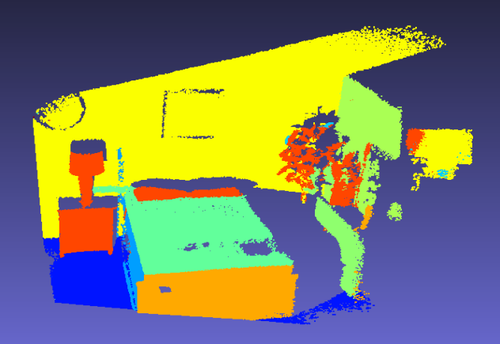} \\
\footnotesize{(a) Input point cloud} & \footnotesize{(b) Output of OPS} \\
\end{tabular}
\end{center}
\vspace{-8pt}
\caption{Sample input and output of OPS. (a) shows a point cloud from the SUN RGB-D dataset \cite{song2015sun}. (b) shows the detected planes after merging.}
\label{figure_sampleOPS}
\vspace{-12pt}
\end{figure}

In this paper, we seek to compare the effectiveness and efficiency of sampling oriented and unoriented points for plane detection in a RANSAC framework. Since sampling one oriented point is sufficient for generating a plane hypotheses, a small number of RANSAC iterations is required to detect the true planes. This, however, requires the computation of relatively accurate surface normals. Since we are interested in planes comprising many points, however, we only have to estimate normals for a small fraction of the points. On the other hand, a sample of three unoriented points can also lead to a plane equation requiring no pre-computations but more RANSAC iterations to draw three inliers from the same plane. This process can be accelerated by sampling the three points in a local neighborhood.  The drawbacks are that it is harder to sample three inliers from the same plane, even locally, and that planes seeded this way are often noisy and may fail the hypothesis verification step.

To enable this comparison, we propose a novel algorithm  called Oriented Points Sampling (OPS), which is based on sparsely sampling points from the point cloud, estimating their normals, using these oriented points to generate plane hypotheses and finally verifying the hypotheses on all points.  The second algorithm is a modification of the work of Biswas and Veloso \cite{biswas2011fast,biswas2012planar}, which is referred to as Fast Sampling Point Filtering (FSPF). FSPF can efficiently detect planes in depth images by sampling points to build and verify plane hypotheses in the vicinity of an original point sampled uniformly from the depth image. We modify the algorithm to sample in spheres around the original point. We evaluate the effectiveness and the computational efficiency of both methods, as well as 3D-KHT algorithm \cite{limberger2015real}, on the SUN RGB-D dataset \cite{song2015sun} that comprises over 10,000 point clouds. Figure \ref{figure_sampleOPS} shows an example from the SUN RGB-D dataset.

The main contributions of this paper are the following:
\begin{itemize}
\item We propose OPS, a fast, RANSAC-based, plane detection method in unorganized point clouds that requires a minimal sample of only one oriented point to generate a hypothesis.
\item We compare OPS on classifying the points of a point cloud according to plane orientation and on plane segmentation with two allternative approaches: an extension of FSPF that samples three unoriented point proposed here and 3D-KHT that operates in octrees. 
\end{itemize}

%% file: relatedwork.tex
There are many different methods of plane detection. These methods can be generally classified into three categories: (i)~point clustering; (ii)~region growing; (iii)~RANSAC-based plane fitting.

Methods based on point clustering rely on similarity measures, such as proximity between points and angle between surface normals. %
Strom et al. \cite{strom2010graph} generated an 8-connected graph from successive laser scans, colored using RGB images, and clustered neighboring points according to differences in color and angle between surface normals. Holz et al. \cite{holz2011real} computed per-point normals and clustered the points according to normal orientation and offset of hypothetical planes going through the oriented points.
Enjarini and Graser \cite{enjarini2012planar} computed the gradient of depth feature on the depth image and performed a voting-based clustering process, followed by region merging.
Pham et al. \cite{pham2016geometrically} initially cluster the input into supervoxels and then perform clustering on the adjacency graph of the supervoxels. Adjacent regions are then classified as belonging to the same or different objects.
Lee and Campbell \cite{lee2016efficient} proposed an efficient surface normal estimation method which discretized the space into grid cells with occupancy information and infered the surface normal direction for each cell utilizing the occupancy information of the neighboring cells. Planes are detected by clustering points with similar normals. %
Dzitsiuk et al. \cite{dzitsiuk2017noising} presented a fast and robust plane detection method for de-noising, 3D reconstruction and hole filling based on Signed Distance Functions (SDFs). The scene is discretized into a set of volumes which contain blocks of voxels with SDF values. Two ways to generate plane candidates in each volume are compared: RANSAC and least squares fitting on the SDF. 

3D-KHT \cite{limberger2015real} is based on a Hough transform and works by clustering approximately co-planar points and by casting votes for these clusters on a spherical accumulator using a trivariate Gaussian kernel. An octree is used to  recursively subdivide the points while a coplanarity test is performed in each node. The recursive subdivision process stops when the points in the node pass the coplanarity test or when there are not enough points in the node.

Methods based on region growing select seed points or regions as the original patches and cluster compatible points with each patch. Poppinga et al. \cite{pop2008fast} proposed a plane fitting algorithm based on region growing followed by a polygonalization algorithm to find a set of convex polygons covering the same area as the set of triangles that is produced by connecting all triplets of points neighboring in the range image. Deschaud and Goulette \cite{deschaud2010fast} presented a fast and accurate algorithm to detect planes in unorganized point clouds using filtered normals and voxel growing. %
Holz and Behnke \cite{holz2013fast} constructed a triangular mesh from the input point cloud and computed local surface normals and curvature estimates. The resulting information was used to segment the range images into planar regions and other geometric primitives.
Arbeiter et al. \cite{arbeiter2014efficient} proposed an efficient normal based region growing method to obtain segments containing points with common geometric properties. Planes are produced after classifying the segments according to point feature descriptors.
Feng et al. \cite{feng2014fast} %
performed agglomerative hierarchical clustering on the point graph to merge nodes belonging to the same plane. The extracted planes were refined using pixel-wise region growing. Monszpart et al. \cite{monszpart2015rapter} proposed a global approach to simultaneously detect a set of planes along with their relations. %
Zermas et al. \cite{zermas2017fast} extracted a set of seed points with low height values which were then used to estimate the initialize the ground surface. Then, points near the initial ground plane were used as seeds to refine the estimate of the ground plane in an iterative process.

RANSAC-based methods address plane detection by sampling points from the point cloud, fitting planar models to them and accepting hypotheses that have accumulated sufficient support.
Oehler et al. \cite{oehler2011efficient} extracted surface elements at multiple resolutions, from coarse to fine. Surface elements that cannot be associated with planes from coarser resolutions are grouped into coplanar clusters using a Hough transform. Connected components are extracted from these clusters and planes are fit using RANSAC. Gallo et al. \cite{gallo2011cc} proposed a modification of the RANSAC algorithm, dubbed CC-RANSAC, that only considers the largest connected components of inliers to evaluate the fitness of a candidate plane. Hulik et al. \cite{hulik2012fast} split the depth map into square tiles and applied RANSAC within each tile.
Then, two seed-fill algorithms are successively applied to group all connected planes in the current tile and across tile borders.
Qian and Ye \cite{qian2014ncc} proposed NCC-RANSAC, which performs a normal coherence check on all points of the inlier patches and removes points whose normal directions are contradictory to that of the fitted plane. The removed points are used to form candidate planes, which are recursively clustered.
Marriott et al. \cite{marriott2017plane} %
extract planar models from depth-data by fitting the data with a piecewise-linear Gaussian mixture regression model and fusing contiguous and coplanar components to generate the final set of planes.

An integral part of our paper is the FSPF algorithm of Biswas and Veloso \cite{biswas2011fast,biswas2012planar}, which groups points in the depth image and  classifies them as belonging to planes or not. To form a plane hypothesis, three points are sampled in a local area, %
while support for the hypothesis is also estimated locally for efficiency. The detected planes are converted into a set of convex polygons, which are  merged across successive depth images. See Section \ref{sec:approach} for more details.

%% file: approach.tex
In this section, we present a method to detect planes of all orientations in a point cloud. We show results on horizontal planes, vertical planes and other planes which are of interest as support surfaces, bounding walls and other objects. We introduce two methods. The first one is Oriented Point Sampling (OPS) which can find all-direction planes in real time. The second one is a fast sampling plane filtering method inspired by \cite{biswas2011fast,biswas2012planar} which we use to compare with OPS in accuracy and speed.

\subsection{Oriented Point Sampling} 
OPS accepts as input a point cloud denoted by $P = \left\{ p_1, p_2, p_3, \cdots, p_N \right\}, p_i \in \mathbb{R}^3$, where $N$ is the number of points. The output is a set of planes $\Pi$. Each plane is stored in a $2 \times 3$ matrix $M$. The first row of $M$ is the plane's centroid and the second row of $M$ is its normal.

Processing begins by estimating the normals of a fraction $\alpha_s$ of the points. On one hand, the availability of oriented points allows the generation of plane hypotheses by sampling just one inlier. On the other hand, accurate normal estimation is computationally expensive and the planes of interest should comprise large numbers of points. Therefore, we only compute the normals of a subset of the points, but use all points for hypotheses verification.

\subsubsection*{Normal Estimation}

\setlength{\textfloatsep}{0pt} 
\begin{algorithm}[!h]
	\SetKwInOut{Input}{Input}
	\SetKwInOut{Output}{Output}
	\caption{OPS}
	\label{algorithm_OPS}
	\Input{An unorganized point cloud $P = \left\{ p_1, p_2, p_3, \cdots, p_N \right\}, p_i \in \mathbb{R}^3$, sampling rate $\alpha_s$, $k$, $p$, distance threshold $\theta_h$, inlier threshold $\theta_N$}
	\Output{The largest horizontal plane model $M$, the inlier set $bestInliers$}
	Randomly sample $\alpha_s  N$ 3D points $P_s = \left\{ p^{(s)}_1, p^{(s)}_2, p^{(s)}_3, \cdots, p^{(s)}_{N_s} \right\}, p^{(s)}_i \in \mathbb{R}^3$\\
	Compute their normals using local SVD on $k$ neighbors \\
	The normals of the sampled points are $\left\{ n^{(s)}_1, n^{(s)}_2, n^{(s)}_3, \cdots, n^{(s)}_{N_s} \right\}$ \\
	$iter \leftarrow 0$\\
	$inNum \leftarrow 0$\\
	$N_{iter} \leftarrow N$\\
	\While{$iter < N_{iter}$} {
		$inliers \leftarrow NULL$\\
		$randInd \leftarrow rand() \% N_s$\\
		\For{each $p^{(s)}_i \in P_s$} {
			$p_r \leftarrow p^{(s)}_i -  p^{(s)}_{randInd}$\\
			$d \leftarrow p_r \cdot N^{(s)}_{randInd}$\\
			\If{$d < \theta_h$} {
				Add $p^{(s)}_i$ to $inliers$\\
			}
		}
		\If{$inliers.size() > \theta_N$} {
			\If {$inliers.size() > inNum$}{
				$inNum \leftarrow inliers.size()$\\
				$bestInliers \leftarrow inliers$\\
				$e \leftarrow 1-\left( \frac{inliers.size()}{N_h} \right)$\\
				$N_{iter} \leftarrow \frac{log\left( 1-p \right)}{log\left(1 - \left( 1 - e \right)\right)}$\\
			}
		}
		$iter \leftarrow iter+1$\\
	}
	\For{each $bestInliers_i \in bestInliers$} {
		$pointsInlier(i,:) \leftarrow bestInliers_i$\\
	}
	$(u,v) \leftarrow SVDCompute(pointsInlier)$\\
	$M(1,:) \leftarrow v(2,:)$\\
	$M(0,:) \leftarrow$ centroid of plane\\
	\Return $M$, $bestInliers$
\end{algorithm}

Given the sampling rate $\alpha_s$, we uniformly sample a subset of the points $P_s = \left\{ p^{(s)}_1, p^{(s)}_2, p^{(s)}_3, \cdots, p^{(s)}_{N_s} \right\}, p^{(s)}_i \in \mathbb{R}^3$, where $N_s$ is the cardinality of $P_s$. We assume that surface normals can be estimated by performing SVD on the neighbors of the reference point, for which we want to estimate the normal.

The current reference point is denoted by $p_i = \left( p_{i,x}, p_{i,y}, p_{i,z} \right)$ and its normal is denoted by $n_i = \left( n_{i,x}, n_{i,y}, n_{i,z} \right)$. The $k$ nearest neighbors of $p_i$ are found using a k-d tree and denoted by $Q_i = \left\{ q_{i1}, q_{i2}, q_{i3}, \cdots, q_{ik} \right\}, q_{ij} \in P$ and $q_{ij} \neq p_{i}$. $k$ is a key parameter for the accuracy of normal estimation. According to \cite{jordan2014quantitative}, we compute the following matrix $M_i$ of each reference point and compute the normal as the eigenvector corresponding to the minimum eigenvalue of $M_i$.

\begin{equation}
M_i = \sum_{q_{ij} \in Q_i} e^{- \frac{\left\| q_{ij} - p_i \right\|_2^2}{2\sigma^2}} \frac{\left( q_{ij} - p_i \right) \left( q_{ij} - p_i \right)^T}{\left\| q_{ij} - p_i \right\|_2^2}
\end{equation}

\noindent This formula includes a weight term $e^{- \frac{\left\| q_{ij} - p_i \right\|_2^2}{2\sigma^2}}$ in order to reduce the effects of neighboring points which are far from the reference point $p_i$.
All vectors connecting neighbors to the reference point are normalized.

\subsubsection*{One-Point RANSAC}

After normal computation, we apply one-point RANSAC to find the largest plane. As opposed to conventional RANSAC-based plane detection, that requires minimal samples of three points, OPS requires only one point with its normal to define a plane. Given the set of sampled points $P_s$, we randomly pick one point with its normal which determine a plane. We compute the distance from all the other sampled points to this plane and count the number of inliers that are within a distance threshold $\theta_h$. We also define a threshold value $\theta_N$ to decide the minimum number of points for a plane to be accepted.

We iterate the above steps to find the plane model with the largest number of inliers. The number of iterations $N_{iter}$ is adaptively determined by $N_{iter} = \frac{log\left( 1 - p \right)}{log\left(1 - \left( 1 - e \right)\right)}$ \cite{hartley04}. $N_{iter}$ is initialized as the number of all the points in the point cloud. $p$ is the probability that at least one random sample is free from outliers and $e$ is the proportion of outliers among all points. $e$ is updated in each iteration as we find more inliers. Since we only need to select one true inlier, the number of iterations is much smaller than alternatives that require sampling three inliers. After we find the largest plane, we re-estimate the normal using SVD on all inliers. OPS is summarized in Algorithm \ref{algorithm_OPS}.

\subsubsection*{Detection of Multiple Planes}
To detect all planes, we apply OPS multiple times. After a plane has been detected, we remove all its inliers from the point cloud and apply OPS on the remaining points until there are not enough points ($\theta_N$) to constitute a plane.

\subsubsection*{Detecting planes in multiple orientations}
To extract basic scene semantics, we divide the detected planes into three groups: vertical, horizontal and other. We have tried two strategies to detect these three groups of planes after sampling some points. (1)~Apply OPS on the sampled points and divide the detected planes into three groups. (2)~Divide the sampled points into three groups based on the estimated normals, and then apply OPS separately on each of the three groups. We chose the second strategy due to its superior performance in our experiments, as shown in Section \ref{sec:experiments}.

\subsection{Fast Sampling Plane Filtering}
In order to compare with OPS, we make some modifications to the Fast Sampling Plane Filtering method \cite{biswas2011fast,biswas2012planar}, which was designed for depth images. Since we are interested in unorganized point clouds, which, unlike RGB-D images, do not take 2D parametrizations, we make the following two modifications to FSPF.
First, while the original FSPF samples neighboring points by adding random integers to the image coordinates of the reference point, we sample neighboring points in a sphere around the reference point. Second, we design a new plane merging method described in Section \ref{sec:merging}. We use FSPF to refer to the algorithm including these modifications in this paper.

\setlength{\textfloatsep}{0pt} 
\begin{algorithm}[!h]
 \SetKwInOut{Input}{Input}
 \SetKwInOut{Output}{Output}
 \SetKwRepeat{Do}{do}{until}
\caption{Fast Sampling Plane Filtering}
\label{algorithm_FSPF}
\Input{An unorganized point cloud $P = \left\{ p_1, p_2, p_3, \cdots, p_N \right\}, p_i \in \mathbb{R}^3$, $N_{max}$, $K_{max}$, $N_{loc}$, $\alpha_{min}$, $\theta_h$, $r_1$, $r_2$}
\Output{A set of planes $\Pi$ and the corresponding inliers}
$\Pi \leftarrow NULL$\\
$n, k \leftarrow 0$\\
\While{$n < N_{max} \bigwedge k < K_{max}$} {
$k \leftarrow k+1$\\
$i = rand(1, N)$\\
$p_{r,0} \leftarrow p_i$\\
$P_1 \leftarrow kdtree.radiusSearch(p_{r,0}, r_1)$\\
$p_{r,1} \leftarrow RandomlyPickPoint(P_1)$\\
$p_{r,2} \leftarrow RandomlyPickPoint(P_1)$\\
$n_\pi = \frac{(p_{r,1} - p_{r,0})\times (p_{r,2} - p_{r,0})}{\left\| (p_{r,1} - p_{r,0})\times (p_{r,2} - p_{r,0}) \right\|_2}$\\
$\hat{\Pi} \leftarrow NULL$\\
$N_{inlier} \leftarrow 0$\\
$P_2 \leftarrow kdtree.radiusSearch(p_{r,0}, r_2)$\\
\For{$j \leftarrow 3 \cdots N_{loc}$} {
$p_j \leftarrow RandomlyPickPoint(P_2)$\\
$e = abs(n_\pi \cdot (p_j - p_{r,0}))$\\
\If{$e < \theta_h$} {
Add $p_j$ to $\hat{\Pi}$\\
$N_{inlier} \leftarrow N_{inlier} + 1$\\
}
}
\If{$N_{inlier} > \alpha_{min}  N_{loc}$} {
\For{each $\hat{\Pi}_i \in \hat{\Pi}$} {
$inlierPoints(i,:) \leftarrow \hat{\Pi}_i$\\
}
$(u,v) \leftarrow SVDCompute(inlierPoints)$\\
$M(1,:) \leftarrow v(2,:)$\\
$M(0,:) \leftarrow$ centroid of plane\\
Add $M$ to $\Pi$\\
Add inlierPoints to $allInliers$
$n \leftarrow n + N_{inlier}$\\
}

}
\Return $\Pi$, $allInliers$

\end{algorithm}

FSPF (Algorithm \ref{algorithm_FSPF}) begins by sampling three points $p_{r,0}, p_{r,1}, p_{r,2}$ from the 3D point cloud. The first point $p_{r,0}$ is selected uniformly in the point cloud, while $p_{r,1}$ and $p_{r,2}$ are sampled within a sphere with a radius $r_1$ around $p_{r,0}$. We use a k-d tree to find the neighbors in the sphere. $p_{r,0}$, $p_{r,1}$ and $p_{r,2}$ determine a plane $\pi$ and we use a cross product to compute its normal. A search sphere with radius $r_2$ is then constructed around point $p_{r,0}$ using the k-d tree and an additional $N_{loc}$ local samples are randomly sampled in the search sphere. The plane fitting error is computed as the distance between each point in the local samples and the plane $\pi$.
Points are considered inliers if their error is less than $\theta_h$, which has the same value as in OPS. If more than $\alpha_{min} N_{loc}$ points in the search sphere are classified as inliers, then all the points in the sphere are classified as inliers of $\pi$ and $\pi$ is recorded as one of the planes found by FSPF. We repeat the above steps for $K_{max}$ iterations or until we find $N_{max}$ inlier points. %

We use FSPF to find all the planes in a point cloud and divide the detected planes into three groups: vertical, horizontal and other. Filtering planes after the initial three points are sampled is ineffective because the normals estimated via the cross product are very noisy. Grouping these planes into vertical, horizontal and other at this stage leads to the detection of numerous planes that are not correctly classified once their normals are re-estimated using all inliers. Filtering planes only after re-estimation is by far the best strategy.

\subsection{Merging Process for Both Methods}\label{sec:merging}
\begin{figure}
\begin{center}
\vspace{4pt}
\begin{tabular}{cc}
\includegraphics[width=.2 \textwidth]{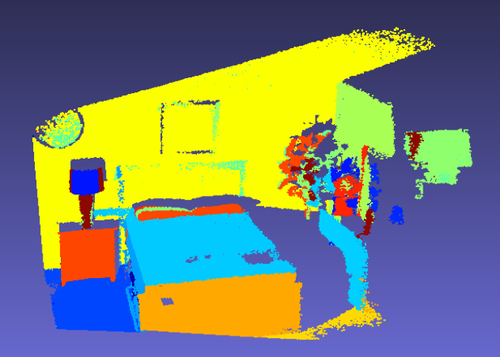} & \includegraphics[width=.2 \textwidth]{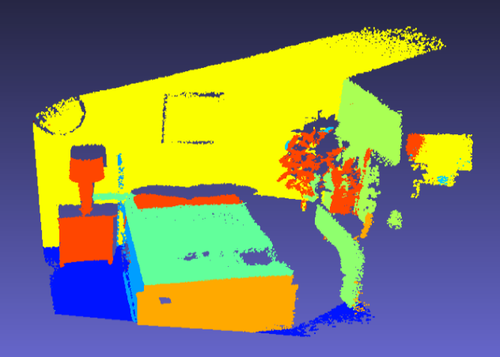} \\
\footnotesize{(a) Unmerged OPS} & \footnotesize{(b) Merged OPS} \\
\includegraphics[width=.2 \textwidth]{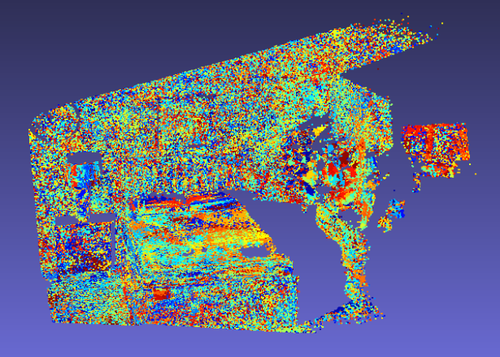} & \includegraphics[width=.2 \textwidth]{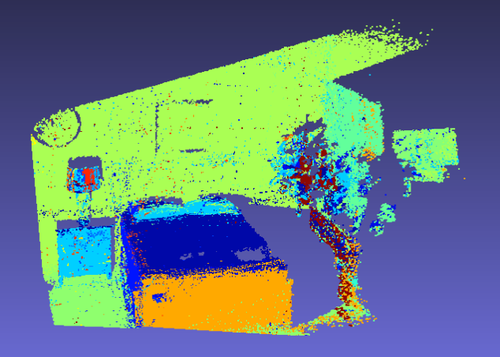} \\
\footnotesize{(c) Unmerged FSPF} & \footnotesize{(d) Merged FSPF} \\
\end{tabular}
\end{center}
\caption{The first row is the output of OPS before and after merging. The second row is the output of FSPF before and after merging.}
\label{figure_merge}
\end{figure}

FSPF returns a lot of small planes because it searches for planes locally. The results of OPS also require some merging, even though it produces larger planes in general. We merge small planes by applying a coplanarity test on each pair of planes. Specifically, we test whether the  (1) angle between the normals of the two planes, (2) the distance between the centroid of the second plane and the first plane, and (3) the distance between the centroid of the first plane and the second plane are below appropriate thresholds.
If a pair of planes passes the coplanarity test, we merge them into one plane and re-estimate the normal of the new plane using SVD on all points. To ensure a fair comparison, we apply the same merging process to both FSPF and OPS. Some results of merging are shown in Figure \ref{figure_merge}.

\setlength{\textfloatsep}{10pt}

%% file: experiment.tex
We perform experiments on the SUN RGB-D dataset \cite{song2015sun} to compare the performance and run time of OPS, FSPF, modified according to Section \ref{sec:approach} and 3D-KHT \cite{limberger2015real}. Using the ground truth planes (see below), we compute the average classification and segmentation accuracy on 10,355 point clouds for all three methods. Classification accuracy refers to assigning each point to the correct plane orientation, while segmentation accuracy refers to assigning each point to the correct plane. We use the Hungarian Algorithm to solve the assignment problem for segmentation accuracy.

\subsection{Datasets and Experimental Setup}

In the experiments, we use the SUN RGB-D dataset \cite{song2015sun} which is captured by four different sensors and contains 10,335 RGB-D images, which \emph{we treat as unorganized point clouds}. The entire dataset is densely annotated and includes 146,617 2D polygons and 64,593 3D bounding boxes. We convert the RGB-D images to point clouds, which are the only inputs to the algorithms. To detect the ground truth planes, we estimate the normals of all the points and iteratively use region growing considering distance and normal similarity to extract dense planes. We use 0.05 m as the distance threshold $\theta_h$ and we use 7 degrees for the threshold of angle between normals. Points on small planes, with under 50 points, are labeled as ``other". Points on horizontal planes are labeled horizontal and points on vertical planes are labeled vertical. 

We run our version of FPSF using the following parameters: the number of local samples $N_{loc}$ is 80, the plane offset error $\theta_h$ is 0.05 m, and the minimum inlier fraction $\alpha_{min}$ to accept a local sample is 0.8. We vary $r_1$, the radius of the sphere for finding $p_{r,1}$ and $p_{r,2}$, and $r_2$, the radius of the search sphere for finding inliers. For 3D-KHT, the minimum number of samples required in a cluster is 30 and the octree level for checking for approximate coplanarity is 5. For OPS, the tolerance to classify a surface normal as vertical or horizontal is 7 degrees, the probability $p$ for adaptively determining RANSAC iterations is 0.99, and the minimum number of points for plane fitting $\theta_N$ is 20. We vary the fraction of points we sample and the number of nearest neighbors for normal estimation. Timing results are reported for single-threaded C++ implementations on an Intel Core i7 7700 processor.

\subsection{Results}

First, we try different values of $r_1$ and $r_2$ for FSPF. We use two configurations: (1) $r_2 = r_1$; (2) $r_2 = 2r_1$. The performance metrics and run time of FSPF are shown in Table \ref{table_FSPF}. The performance improves as $r_1$ increases. FSPF performs the best when $r_1$ is 0.1 m and $r_2 = 2r_1$.

\begin{table}[!h]
	\caption{Classification Accuracy and run time of FSPF.}
	\label{table_FSPF}
	\vspace{-12pt}
	\begin{center}
		\begin{tabular}{|c|c|c|c|c|c|}
			\hline
			$r_1$ & $r_2$ & Accuracy & FSPF Time & Merge Time & Total Time \\
			\hline
			0.03 & 0.03 & 61.34\% & 0.0974 sec & 0.1895 sec & 0.2997 sec \\
			\hline
			0.05 & 0.05 & 68.07\% & 0.1102 sec & 0.2103 sec & 0.3003 sec \\
			\hline
			0.07 & 0.07 & 69.68\% & 0.2106 sec & 0.1937 sec & 0.4043 sec \\
			\hline
			0.1 & 0.1 & 71.14\% & 0.2479 sec & 0.2145 sec & 0.4642 sec \\
			\hline
			0.3 & 0.3 & 73.12\% & 2.6423 sec & 0.3293 sec & 2.9716 sec \\
			\hline
			0.03 & 0.06 & 69.01\% & 0.1139 sec & 0.2415 sec & 0.3554 sec \\
			\hline
			0.05 & 0.1 & 71.11\% & 0.1973 sec & 0.2487 sec & 0.4460 sec \\
			\hline
			\textbf{0.07} & \textbf{0.14} & \textbf{72.50\%} & \textbf{0.3441 sec} & \textbf{0.2785 sec} & \textbf{0.6226 sec} \\
			\hline
			0.1 & 0.2 & 73.61\% & 0.7277 sec & 0.3180 sec & 1.0457 sec \\
			\hline
			0.3 & 0.6 & 66.26\% & 2.7624 sec & 0.4960 sec & 3.2584 sec \\
			\hline
		\end{tabular}
	\end{center}
\end{table}

The run time of FSPF consists of plane detection and merging. The former ranges from 0.1 to 0.7 seconds. FSPF plane detection time becomes longer as $r_1$ increases. When $r_1$ becomes larger than 0.3, the detection is very slow. Since our focus is on real-time plane detection, we only show results for $r_1 \leq 0.3 m$. The merging time ranges from 0.2 to 0.5 seconds. FSPF detects more planes when $r_1$ and $r_2$ become larger. Merging takes longer when $r_1$ increases, since there are more planes to be merged. 

Second, we evaluate the accuracy and run time of 3D-KHT, as shown in Table \ref{table_3dkht}. 3D-KHT uses a spherical accumulator to cast votes based on Hough transform. We tried different number of cells in the angular $\phi$ and radius $r$ direction. The more cells it uses, the more time it costs and the more accurate planes it detects. %

\begin{table}[!h]
    \caption{Classification Accuracy and run time of 3D-KHT.}
    \label{table_3dkht}
    \vspace{-12pt}
    \begin{center}
        \begin{tabular}{|c|c|c|c|}
            \hline
			Angular bins & Radial bins & Accuracy & Run Time \\
			\hline
			30 & 300 & 67.08\% & 0.0837 sec \\
			\hline
			40 & 400 & 67.86\% & 0.1448 sec \\
			\hline
			60 & 600 & 69.53\% & 0.3674 sec \\
			\hline
			\textbf{80} & \textbf{800} & \textbf{71.01\%} & \textbf{0.7941 sec} \\
			\hline
        \end{tabular}
    \end{center}
\end{table}

\begin{table}[!h]
    \vspace{6pt}
	\caption{Classification Accuracy and run time of OPS. The columns show the sampling rate, the number of nearest neighbors, average accuracy, run time for plane detection and total time, including merging.}
	\label{table_OPS}
	\vspace{-12pt}
	\begin{center}
		\begin{tabular}{|c|c|c|c|c|c|}
			\hline
			Sampl. rate & NN & Accuracy & OPS Time & Total Time\\
			\hline
			0.3\% & 10 & 71.52\% & 0.0244 sec & 0.0359 sec \\
			\hline
			0.3\% & 20 & 72.18\% & 0.0347 sec & 0.0460 sec \\
			\hline
			0.3\% & 30 & 72.29\% & 0.0458 sec & 0.0565 sec \\
			\hline
			0.5\% & 10 & 77.11\% & 0.0452 sec & 0.0625 sec \\
			\hline
			0.5\% & 20 & 77.48\% & 0.0644 sec & 0.0803 sec \\
			\hline
			0.5\% & 30 & 77.59\% & 0.0833 sec & 0.0987 sec \\
			\hline
			0.7\% & 10 & 80.04\% & 0.0656 sec & 0.0853 sec \\
			\hline
			0.7\% & 20 & 80.28\% & 0.0906 sec & 0.1093 sec \\
			\hline
			0.7\% & 30 & 80.34\% & 0.1184 sec & 0.1363 sec \\
			\hline
			1\% & 10 & 82.55\% & 0.1008 sec & 0.1233 sec \\
			\hline
			1\% & 20 & 82.66\% & 0.1375 sec & 0.1587 sec \\
			\hline
			1\% & 30 & 82.71\% & 0.1759 sec & 0.1967 sec \\
			\hline
			3\% & 10 & 86.91\% & 0.4092 sec & 0.4377 sec \\
			\hline
			3\% & 20 & 87.23\% & 0.5096 sec & 0.5366 sec \\
			\hline
			\textbf{3\%} & \textbf{30} & \textbf{87.39\%} & \textbf{0.6188 sec} & \textbf{0.6455 sec} \\
			\hline
		\end{tabular}
	\end{center}
	\vspace{-8pt}
\end{table}

Last, we evaluate the effects of different sampling rates and different numbers of nearest neighbors on the performance of OPS. The experimental results are shown in Table \ref{table_OPS}. The average classification accuracy of OPS improves as the sampling rate increases. Using a larger number of nearest neighbors for normal estimation improves the results of OPS at the cost of longer run time. However, when the sampling rate is more than 0.5\% even 10 nearest neighbors are sufficient for OPS to achieve high accuracy. As mentioned in Section \ref{sec:approach}, we tried the first strategy for detecting planes in multiple orientations. The average accuracy is 74.25\%, 75.04\% and 75.89\% when we estimate normals for 3\% of the points and the number of nearest neighbors is 10, 20 and 30. This is worse than the last three rows of Table \ref{table_OPS}.

\begin{figure}
\vspace{4pt}
\begin{center}
\begin{tabular}{cc}
\includegraphics[width=.22 \textwidth]{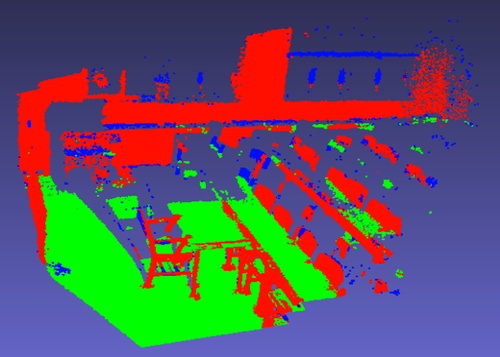} & \includegraphics[width=.22   \textwidth]{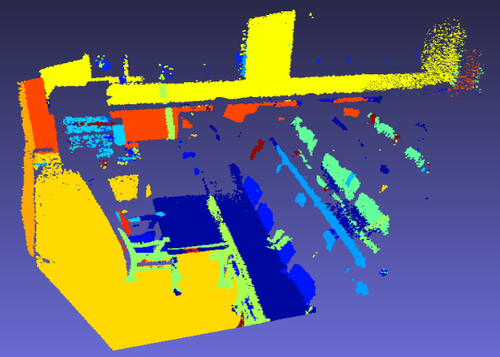} \\
\footnotesize{($a$) Ground Truth for 3 classes} & \footnotesize{($b$) Ground Truth} \\
\includegraphics[width=.22 \textwidth]{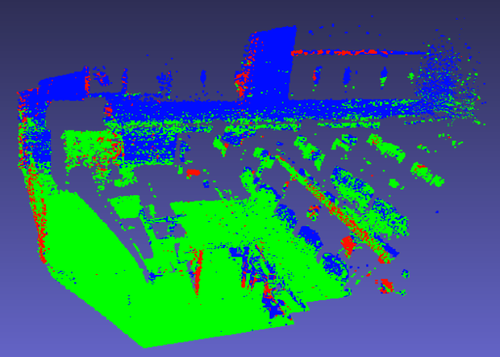} & \includegraphics[width=.22 \textwidth]{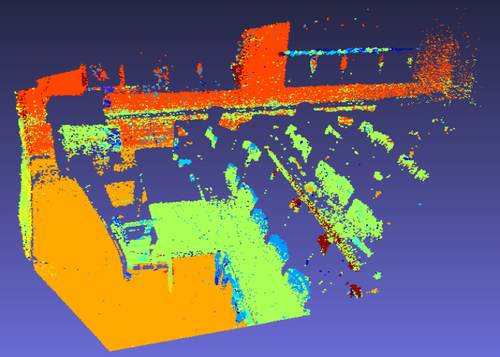} \\
\footnotesize{($c$) FSPF for 3 classes} & \footnotesize{($d$) FSPF} \\
\includegraphics[width=.22 \textwidth]{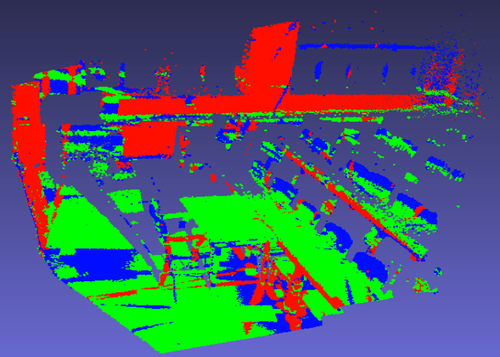} & \includegraphics[width=.22 \textwidth]{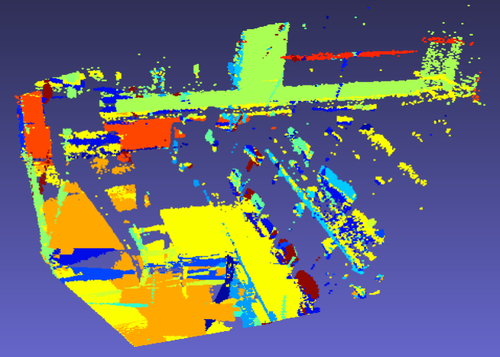} \\
\footnotesize{($e$) 3D-KHT for 3 classes} & \footnotesize{($f$) 3D-KHT} \\
\includegraphics[width=.22 \textwidth]{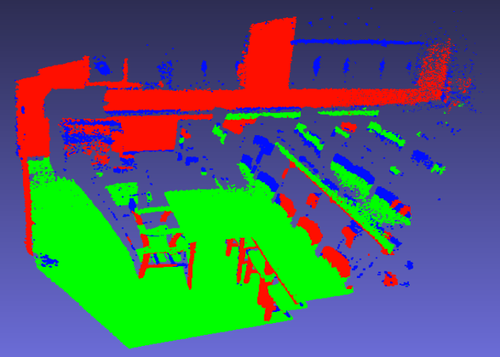} & \includegraphics[width=.22 \textwidth]{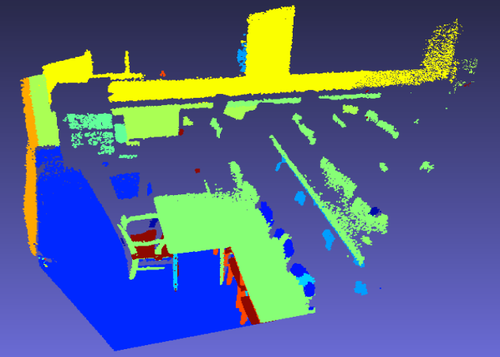} \\
\footnotesize{($g$) OPS for 3 classes} & \footnotesize{($h$) OPS} \\
\end{tabular}
\end{center}
\vspace{-5pt}
\caption{Some results of FSPF, 3D-KHT and OPS. The left and right columns show classification and segmentation results, respectively.}
\label{figure_output}
\end{figure}

OPS is better than FSPF with any $r_1$ in classification accuracy when the sampling rate is between 0.5\% and 3\%. When $r_1 = 0.1, r_2 = 2r_1$, FSPF achieves the highest accuracy with a run time of about 1 second. In addition to higher accuracy, OPS is also faster than FSPF when the sampling rate is between 0.5\% and 3\%. The merging process of OPS is also much faster than that of FSPF because OPS detects larger planes than FSPF. On average, in each point cloud OPS detects 34.62 planes before merging and 27.59 planes after merging. The corresponding numbers for FSPF are 386.92 and 29.44. The ground truth contains on average 25.38 planes. %
The left column of Fig. \ref{figure_output} shows the results of classifying the points according to the three possible orientations and the right column shows results where each plane is marked in a pseudo-random color.

We pick the best parameter settings of the three methods and evaluate segmentation accuracy, as shown in Table \ref{table_segmen}. The best parameter setttings are: (1) $r_1 = 0.07, r_2 = 0.14$ for FSPF; (2) 80 angular bins and 800 radius bins for 3D-KHT; (3) 3\% sampling rate and 30 nearest neighbors for OPS. OPS is able to assign points to the correct plane with much higher accuracy. Note that the ground truth was generated by region growing which different than all three algorithms.
\begin{table}[!h]
    \vspace{0pt}
    \caption{Segmentation Accuracy of FSPF, 3D-KHT and OPS.}
    \label{table_segmen}
    \vspace{-12pt}
    \begin{center}
        \begin{tabular}{|c|c|c|}
            \hline
			FSPF & 3D-KHT & OPS \\
			\hline
			74.72\% & 62.26\% & 82.99\% \\
			\hline
        \end{tabular}
    \end{center}
    \vspace{-16pt}
\end{table}

\setlength{\textfloatsep}{10pt}

%% file: conclusion.tex
In this paper, we introduced a new sampling-based method for detecting plane in point clouds and compared it with two other approaches. We introduced OPS which is based on sparsely sampling points from the point cloud, estimating their normals and then using these oriented points to generate plane hypotheses for verification. We modified FSPF, which was originally designed for depth images, to be applicable to unorganized point clouds. We also used the 3D-KHT algorithm which relies on an octree to cluster approximately coplanar points. We experimentally evaluated the accuracy and computational efficiency of all methods on the large scale SUN RGB-D dataset. Our conclusions are that OPS is superior in both speed and accuracy.
These experiments provide evidence that may help resolve the debate on whether investing computational resources for normal estimation is beneficial for plane detection. While estimating the normals of all points using more than 10 nearest neighbors and a large number of queries to a k-d tree is too costly in terms of computation, it is sufficient to compute normals for only a small fraction of the points, as low as 0.3\% in these experiments. The availability of these privileged, oriented points allows the use of a one-point RANSAC scheme which can detect almost all planes in the scene. %